\documentclass[10pt,journal,compsoc]{IEEEtran}
\pdfoutput=1

\usepackage{graphicx}

\usepackage{tikz}
\usepackage{comment}
\usepackage{amsmath,amssymb} 
\usepackage{color}
\usepackage{colortbl}
\usepackage{booktabs}
\usepackage{multirow}
\usepackage{wrapfig}
\usepackage{graphicx}
\usepackage{subfigure}
\usepackage{pifont}
\usepackage{ragged2e}
\usepackage{url}

\usepackage{xcolor}
\usepackage{orcidlink}
\usepackage{marvosym}

\usepackage{algorithm}
\usepackage{algorithmic}
\definecolor{tabfirst}{rgb}{1, 0.7, 0.7}
\definecolor{tabsecond}{rgb}{1, 0.85, 0.7}
\definecolor{tabthird}{rgb}{1, 1, 0.7}

\usepackage{amsmath,amsfonts,bm}
\def\1{\bm{1}}

\DeclareMathOperator*{\argmin}{arg\,min}

\def\mI{{\bm{I}}}

\DeclareMathAlphabet{\mathsfit}{\encodingdefault}{\sfdefault}{m}{sl}
\SetMathAlphabet{\mathsfit}{bold}{\encodingdefault}{\sfdefault}{bx}{n}

\def\gU{{\mathcal{U}}}

\newcommand{\eg}{\textit{e.g.~}}

\ifCLASSOPTIONcompsoc
  \usepackage[nocompress]{cite}
\else
  \usepackage{cite}
\fi

\begin{document}
%
\title{ViewCrafter: Taming Video Diffusion Models for High-fidelity Novel View Synthesis
} 

%
%

\author{Wangbo Yu*,
        Jinbo Xing*,
        Li Yuan*,
        Wenbo Hu$^\dagger$,
        Xiaoyu Li,
        Zhipeng Huang,\\
        Xiangjun Gao,
        Tien-Tsin Wong,
        Ying Shan,
        and
        Yonghong Tian$^\dagger$,~\IEEEmembership{Fellow,~IEEE}

\IEEEcompsocitemizethanks{\IEEEcompsocthanksitem *: Equal contribution. $\dagger$: Corresponding authors. Wangbo Yu, Li Yuan, Zhipeng Huang, Yonghong Tian are with Peking University and Peng Cheng Laboratory. Jinbo Xing is with The Chinese University of Hong Kong. Wenbo Hu, Xiaoyu Li are with Tencent AI Lab. Ying Shan is with ARC Lab, Tencent PCG and Tencent AI Lab. Xiangjun Gao is with Hong Kong University of Science and Technology. Tien-Tsin Wong is with Monash University.
}
}

%
%

\markboth{ }%
{Shell \MakeLowercase{\textit{et al.}}: Bare Demo of IEEEtran.cls for Computer Society Journals}
%



\IEEEtitleabstractindextext{%

\begin{abstract}
\justifying 
Despite recent advancements in neural 3D reconstruction, the dependence on dense multi-view captures restricts their broader applicability. In this work, we propose \textbf{ViewCrafter}, a novel method for synthesizing high-fidelity novel views of generic scenes from single or sparse images with the prior of video diffusion model. Our method takes advantage of the powerful generation capabilities of video diffusion model and the coarse 3D clues offered by point-based representation to generate high-quality video frames with precise camera pose control. To further enlarge the generation range of novel views, we tailored an iterative view synthesis strategy together with a camera trajectory planning algorithm to progressively extend the 3D clues and the areas covered by the novel views. With ViewCrafter, we can facilitate various applications, such as immersive experiences with real-time rendering by efficiently optimizing a 3D-GS representation using the reconstructed 3D points and the generated novel views, and scene-level text-to-3D generation for more imaginative content creation. Extensive experiments on diverse datasets demonstrate the strong generalization capability and superior performance of our method in synthesizing high-fidelity and consistent novel views. Our project webpage and code are available at \url{https://drexubery.github.io/ViewCrafter/}.

\end{abstract}

\begin{IEEEkeywords}
Novel View Synthesis, Video Diffusion Models, 3D Scene Generation.
\end{IEEEkeywords}}

\maketitle

\IEEEdisplaynontitleabstractindextext

%
\IEEEpeerreviewmaketitle

\IEEEraisesectionheading{\section{Introduction}\label{sec:intro}}

\IEEEPARstart{N}{ovel} view synthesis plays a crucial role in computer vision and graphics for creating immersive experiences in games, mixed reality, and visual effects.
Despite the significant success of 3D neural reconstruction techniques such as NeRF\cite{mildenhall2020nerf} and 3D-GS\cite{kerbl20233dgs}, their dependence on dense multi-view observations restricts their broader applicability in situations where only limited views are available.

A more desirable problem scenario in practice involves synthesizing novel views of generic scenes from sparse observations or even a single image.
This task is considerably challenging as it necessitates a comprehensive understanding of the 3D world, including structures, appearance, semantics, and occlusions.  
Early researches\cite{wiles2020synsin,rombach2021geometryfree,rockwell2021pixelsynth,park2024bridging,zhou2018real10k,han2022single,single_view_mpi,yu2021pixelnerf} focused on training regression-based models to synthesize novel views from sparse or single input. However, due to their limited representation capabilities, these methods are mostly category-specific and only handle certain domains such as indoor scenes. 
Recent advancements in powerful diffusion models have made zero-shot novel view synthesis\cite{liu2023zero,zeronvs,wang2023motionctrl} from single view approachable. 
Nevertheless, these methods are either restricted to handling object-level images or lack precise control of the camera pose due to their dependency on high-level pose prompts to guide the view synthesis process.
Some works\cite{chung2023luciddreamer,shriram2024realmdreamer} also attempt to synthesize novel views from a single image using depth-based warping and diffusion-based inpainting.
Yet, these methods often produce inconsistent content in occlusion regions due to the per-view inpainting mechanism.

In this work, we focus on high-fidelity novel view synthesis of \emph{generic scenes} from single or sparse images, maintaining \emph{precise control of the camera pose} and \emph{consistency} of the generated novel views.
To achieve this, we investigate leveraging the generative capabilities of video diffusion models alongside the explicit 3D information provided by point cloud representations. On one hand, 
video diffusion models\cite{pku_yuan,blattmann2023svd,xing2023dynamicrafter}, trained on web-scale video datasets, develop a reasonable understanding of the world, which facilitates the generation of plausible video content from a single image or text prompt.
However, they lack the underlying 3D information of the scene and fall short in achieving precise camera view control.
%
On the other hand, recent dense stereo methods \cite{wang2024dust3r,mast3r_arxiv24} have made fast point cloud reconstruction from single or sparse images accessible. Point cloud representation provides valuable coarse 3D scene information, enabling precise pose control for free-view rendering. 
Yet, due to its poor representation capability and the limited 3D cues offered by extremely sparse reference images, it suffers from occlusions, missing areas, and geometry distortion, hindering its utility in novel view synthesis.
With these in mind, we propose integrating the generative power of video diffusion models with the coarse 3D prior provided by point-based representation, aiming to facilitate higher-fidelity novel view synthesis of generic scenes.

Our method, \textbf{ViewCrafter}, accomplishes novel view synthesis by a \emph{point-conditioned video diffusion model} that generates high-fidelity and consistent videos under a novel view trajectory, conditioned on corresponding frames rendered from point cloud reconstructed from single or sparse images.
Leveraging the explicit 3D information from the point cloud and the generative capabilities of video diffusion models, our method enables precise control of 6 DoF camera poses and generates high-fidelity, consistent novel views.
%
%
Furthermore, video diffusion models face challenges in generating long videos because of unacceptably increased memory and computation costs.
To tackle this challenge, we propose an \emph{iterative view synthesis} strategy along with a \emph{content-adaptive camera trajectory planning} algorithm to progressively extend the reconstructed point cloud and the areas covered by the novel views.
Starting from the initial point cloud derived from the input image(s), we first employ the camera trajectory planning algorithm to predict a content-adaptive camera pose sequence from the current point cloud to effectively reveal occlusions.
Next, we render the point cloud according to the predicted pose sequence and synthesize novel views by ViewCrafter with the conditions of the rendered point cloud. Subsequently, the point cloud is updated from the synthesized novel views to extend the global point cloud representation.
Through iteratively conducting these steps, we can ultimately obtain high-fidelity novel views that cover a large view range and an extended point cloud.

In addition to novel view synthesis, we explore several applications of our method. For instance, we can efficiently optimize a 3D-GS representation based on the constructed point cloud and the synthesized novel views within minutes, enabling real-time rendering for immersive experiences. 
Furthermore, our method shows the potential to adapt to scene-level text-to-3D generation, which can foster more imaginative artistic creations.

We extensively evaluate our method for zero-shot novel view synthesis and sparse view 3D-GS reconstruction on various datasets, including Tanks-and-Temples\cite{Knapitsch2017tnt}, RealEstate10K\cite{zhou2018real10k}, and CO3D\cite{reizenstein2021co3d}.
For zero-shot novel view synthesis, our method outperforms the baselines in both image quality and pose accuracy metrics. This demonstrates its superior ability to synthesize high-fidelity novel views and achieve precise pose control. In 3D-GS reconstruction, our approach consistently surpasses previous state-of-the-art. This further validates its effectiveness in scene reconstruction from sparse views.
%

Our contributions can be summarized as follows:
\begin{itemize}
    \item We propose \textbf{ViewCrafter}, a novel view synthesis framework tailored for synthesizing high-fidelity novel view sequences of generic scenes from single or sparse images while maintaining precise control of camera poses.
    \item We present an iterative view synthesis strategy in conjunction with a content-adaptive camera trajectory planning algorithm to progressively expand the covered areas of novel views and the reconstructed point cloud, enabling long-range and large-area novel view synthesis.
    \item Our method achieves superior performance on various challenging datasets in terms of both the quality of synthesized novel views and the accuracy of camera pose control.
    It facilitates various applications beyond novel view synthesis, such as real-time rendering for immersive experiences by efficiently optimizing a 3D-GS representation from our results, and scene-level text-to-3D generation for more imaginative artistic creations.
\end{itemize}

\begin{figure*}[t]
  \centering
  \includegraphics[width=1.\textwidth]{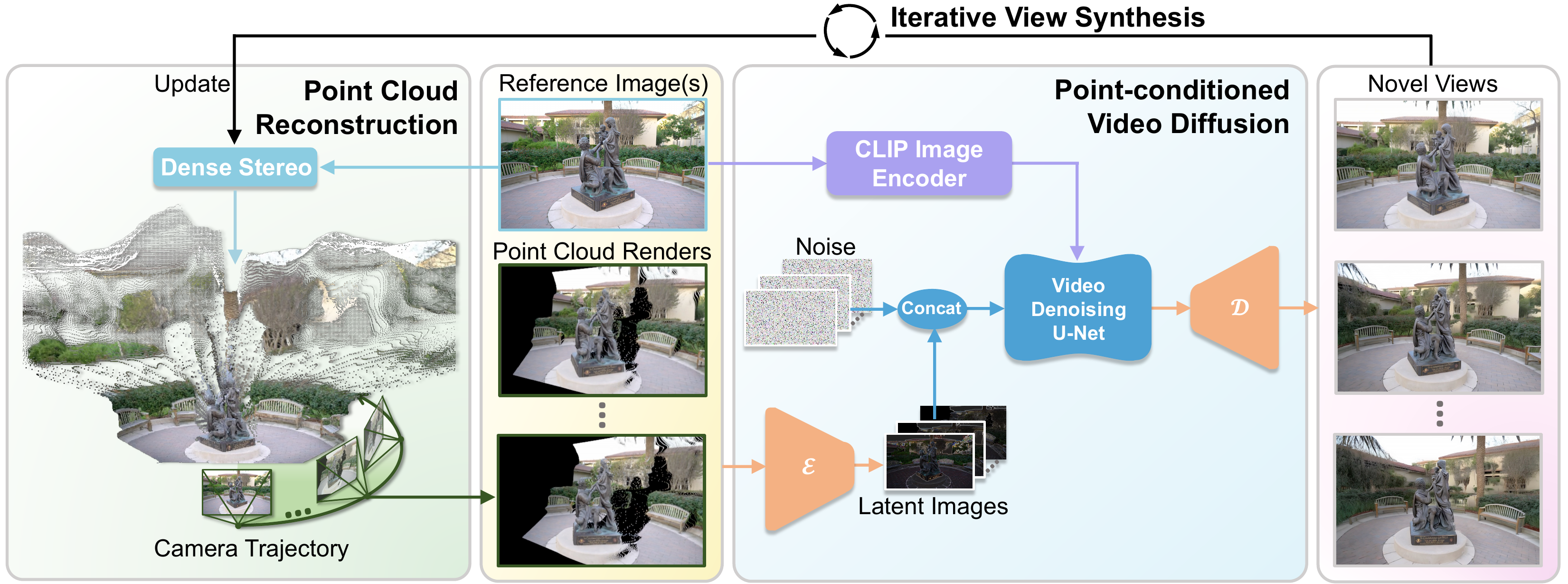}
  \caption{\textbf{Overview of our ViewCrafter.} Given a single reference image or sparse image sets, we first build its point cloud representation using a dense stereo model, which enables accurately moving cameras for free-view rendering. Subsequently, to address the large missing regions, geometric distortions, and point cloud artifacts exhibited in the point cloud render results, we train a point-conditioned video diffusion model to serve as an enhanced renderer, facilitating the generation of high-fidelity and consistent novel views based on the coarse point cloud renders. To achieve long-range novel view synthesis, we adopt an iterative view synthesis strategy that involves iteratively moving cameras, generating novel views, and updating the point cloud, which enables a more complete point cloud reconstruction and benefits downstream tasks such as 3D-GS optimization.
}
\label{fig:pipeline1}
\end{figure*}
\section{Related Work}

\label{sec:review}
\subsection{Regression-based Novel View Synthesis}

Regression-based methods aim to train a feed-forward model to generate novel views from sparse or single image inputs.
This is often achieved using CNN/Transformer-based\cite{vaswani2017attention} architectures to establish a 3D representation of the input image(s). 
For instance, several works\cite{trevithick2023lp3d,yu2023nofa} have applied this idea to specific modalities, such as human faces, by generating tri-plane representations for novel view synthesis. 
LRM\cite{honglrm} extends this strategy to generic objects, while other methods like\cite{zhou2018real10k,han2022single,single_view_mpi} adopt the multi-plane representation, and PixelNeRF\cite{yu2021pixelnerf} employs NeRF\cite{mildenhall2020nerf} as 3D representation for novel view synthesis. 
Inspired by the success of 3D-GS\cite{kerbl20233dgs}, recent approaches such as PixelSplat\cite{charatan2024pixelsplat} and MVSplat\cite{chen2024mvsplat} explore training regression-based models to produce 3D Gaussian representations for real-time rendering capabilities.
Additionally, some methods like\cite{wiles2020synsin,rombach2021geometryfree,rockwell2021pixelsynth,park2024bridging} combine monocular depth estimation and image inpainting modules in a unified framework for novel view synthesis.
However, these methods are limited to category-specific domains, such as objects and indoor scenes, and are prone to artifacts due to their limited model representation capabilities. In contrast, our method can synthesize high-fidelity novel views of generic scenes.

\subsection{Diffusion-based Novel View Synthesis}
The rapid advancement of diffusion models\cite{ho2020denoising,song2021denoising,rombach2022high} have demonstrated exceptional proficiency in synthesizing high-quality images and shows the potential to be adapted in synthesizing novel views from single or sparse inputs. 
While some optimization-based approaches\cite{yu2023hifi, Tang_2023_makeit,sun2023dreamcraft3d} directly train a 3D representation under the supervision of text-to-image (T2I) diffusion models\cite{rombach2022high}, they require scene-specific optimization, which compromising their generalization capabilities. 
To address this, GeNVS\cite{chan2023genvs} proposes a generalized novel view synthesis framework by training a 3D feature-conditioned diffusion model on a large-scale multiview dataset\cite{reizenstein2021co3d}. Similarly, Zero-1-to-3\cite{liu2023zero} develops camera pose-conditioned diffusion models trained on synthetic datasets\cite{deitke2023objaverse,chang2015shapenet}, enabling novel view synthesis from more diverse inputs. However, these models are either category-specific\cite{chan2023genvs,3dim} or limited to handling toy-level objects with simple backgrounds.
Recently, ZeroNVS\cite{zeronvs} improves the generation capability of  Zero-1-to-3\cite{liu2023zero} by training it on a mixed dataset containing both synthetic\cite{deitke2023objaverse} and real data\cite{zhou2018real10k,reizenstein21co3d,yu2023mvimgnet}, enabling zero-shot novel view synthesis of generic scenes from a single input image. Nonetheless, it still struggles to synthesize consistent novel views and lacks precise pose control, as it treats camera pose conditions as high-level text embeddings. 
Reconfusion\cite{wu2024reconfusion} proposes a PixelNeRF\cite{yu2021pixelnerf} feature-conditioned diffusion model to achieve relatively accurate pose control in novel view synthesis. However, it fails to synthesize consistent novel views due to its inability to model the correlations among sampled views. Additionally, it requires multiple images as input and cannot process a single image.
Several works\cite{zhang2024text2nerf,chung2023luciddreamer,shriram2024realmdreamer} utilize depth-based warping to synthesize novel views and employ a pre-trained T2I diffusion model\cite{Rombach2022sd2} to refine the warped images. However, the novel views generated by these methods often suffer from artifacts and unrealistic contexts in the inpainted regions, which limits their applicability.

\subsection{Conditional Video Diffusion Models}
As generative models that produce content from diverse input modalities evolve rapidly, enhancing user control over generation has garnered significant interest. ControlNet\cite{zhang2023adding}, T2I-adapter\cite{mou2024t2i}, and GLIGEN\cite{li2023gligen} pioneered the introduction of condition signals for T2I generation. Similar strategies have also been employed in video generation, enabling controls like RGB images\cite{blattmann2023svd,xing2023dynamicrafter,xing2024tooncrafter}, depth\cite{xing2024make,esser2023structure}, trajectory\cite{yin2023dragnuwa,niu2024mofa}, and semantic maps\cite{peruzzo2024vase}. However, camera motion control has received comparatively less attention.
AnimateDiff\cite{guo2023animatediff} and SVD\cite{blattmann2023svd} investigate class-conditioned video generation, grouping camera movements and utilizing LoRA\cite{hu2021lora} modules to create specific camera motions. MotionCtrl\cite{wang2023motionctrl} improves control by using camera extrinsic matrices as conditioning signals. Although effective for simple trajectories, their dependence on 1D numeric values leads to imprecise control in complex real-world situations. 
MultiDiff\cite{muller2024multidiff} leverage depth-based warping to produce warped images, and condition the video diffusion model on the warped images to provide explicit 3D prior. Nevertheless, it trains the video diffusion model on class-specific datasets\cite{zhou2018real10k,dai2017scannet}, thereby lacking the generalization ability to handle generic scenes. Recently,  CamCo\cite{xu2024camco} and CameraCtrl\cite{he2024cameractrl} introduced Plücker coordinates\cite{sitzmann2021light} in video diffusion models for camera motion control. Nevertheless, these methods still cannot precisely control the camera motion due to the complicated mapping from numeric camera parameters to videos. In this paper, we propose to leverage explicit point cloud representations for precise camera control in video generation, thereby fulfilling our needs for consistent and accurate novel view synthesis. 



\section{Method}
In the following, we start with a brief introduction to video diffusion models in Section.~\ref{subsec:prelimiary}, followed by an explanation of the point cloud reconstruction pipeline in Section.~\ref{subsec:pc} and an illustration of the point-conditioned video diffusion model in Section.~\ref{subsec:vid}. Subsequently, we elaborate on the iterative view synthesis and camera trajectory planning strategy in Section.~\ref{subsec:NBV}, and demonstrate how to apply our approach for efficient 3D-GS optimization and text-to-3D generation in Section.~\ref{subsec:3D-GS}.

\subsection{Preliminary: Video Diffusion Models}
\label{subsec:prelimiary}
A diffusion model\cite{song2021denoising} consists of two primary components: a forward process $q$ and a reverse process $p_\theta$. The forward process initiates with clean data $\bm{x}_0 \sim q_0(\bm{x}_0)$ and gradually introduces noise to $\bm{x}_0$, creating a noisy state across different time steps. This is mathematically represented as $\bm{x}_t = \alpha_t\bm{x}_0+\sigma_t\epsilon$, where $\epsilon\sim\mathcal{N}(\bm{0},\mI)$. The hyper-parameters $\alpha_t$ and $\sigma_t$ satisfy the constraint $\alpha^2_t + \sigma^2_t = 1$. The reverse process $p_\theta$ focuses on removing noise from the clean data utilizing a noise predictor $\epsilon_\theta$, which is optimized by the objective:
\begin{equation}
\min _{\theta} = \mathbb{E}_{t\sim\gU(0,1),\epsilon\sim\mathcal{N}(\bm{0},\mI)}[\|\epsilon_\theta(\bm{x}_t,t) - \epsilon\|_2^2].
\end{equation}

In diffusion-based video generation, Latent Diffusion Models (LDMs)\cite{metzer2022latent} are frequently employed to mitigate the computational cost. In LDMs, the video data $\bm{x} \in \mathbb{R}^{L\times 3\times H\times W}$ are encoded into the latent space using a pre-trained VAE encoder frame-by-frame, expressed as $\bm{z}=\mathcal{E}(\bm{x})$, $\bm{z} \in \mathbb{R}^{L\times C\times h\times w}$. Then, both the forward process $q$ and the reverse process $p_{\theta}$ are performed in the latent space. The final generated videos are obtained through the VAE decoder $\hat{\bm{x}}=\mathcal{D}(\bm{z})$.
In this work, we build our model based on an open-sourced Image-to-Video (I2V) diffusion model DynamiCrafter\cite{xing2023dynamicrafter}, which is capable of creating dynamic videos from a single input image. This aligns naturally with our goal of synthesizing novel views from sparse or single inputs.

\subsection{Point Cloud Reconstruction from Single or Sparse Images}
\label{subsec:pc}

To achieve accurate pose control in our novel view synthesis framework, we first establish the point cloud representation from the reference image(s).
Specifically, 
we employ a dense stereo model, \eg DUSt3R\cite{wang2024dust3r}, to reconstruct the point cloud and estimate camera parameters simultaneously.
It takes a pair of RGB images $\mathbf{I}^0, \mathbf{I}^1 \in \mathbb{R}^{H\times W \times 3}$ as input and generates corresponding point maps $\mathbf{O}^{0,0},\mathbf{O}^{1,0} \in \mathbb{R}^{H\times W \times 3}$ along with their respective confidence maps $\mathbf{D}^{0,0},\mathbf{D}^{1,0} \in \mathbb{R}^{H\times W}$, reflecting the level of confidence in their accuracy.
The subscript of $\mathbf{O}^{0,0}, \mathbf{O}^{1,0}$ denote that they are expressed in the same camera coordinate system of the anchor view $\mathbf{I}^0$. 
To recover the camera's intrinsic parameters, it is assumed that the principal point is centered and the pixels are square. Consequently, only the focal length $\bm{f}_0^*$ remains unknown, which can be solved through a few optimization steps using Weiszfeld algorithm\cite{plastria2011weiszfeld}:
\begin{equation}
  \bm{f}_0^* = \argmin_{\bm{f}_0} \sum_{u=0}^{W} \sum_{v=0}^{H} \mathbf{D}^{0,0}_{u,v} \left\Vert (u',v') - \bm{f}_0 \frac{(\mathbf{O}^{0,0}_{u,v,0}, \mathbf{O}^{0,0}_{u,v,1})}{\mathbf{O}^{0,0}_{u,v,2}} \right\Vert,
\end{equation}
where $u'=u-\frac{W}{2}$ and $v'=v-\frac{H}{2}$. 
In the case where only a single input image is available, we duplicate the input image to create a paired input and then estimate its point map and camera intrinsic. When there are more than two input images, it can also perform global point map alignment with a few optimization iterations.

The colored point cloud, which provides coarse 3D information of the scene, can be obtained by integrating the point maps with their corresponding RGB images. However, the limited representation capabilities of the point cloud and the insufficient 3D cues provided by sparse or single inputs can result in significant missing regions, occlusions, and artifacts in the reconstructed point cloud, leading to low-quality render results. Therefore, we propose incorporating video diffusion models to achieve high-fidelity novel view synthesis based on the imperfect point cloud.

\subsection{Rendering High-fidelity Novel Views with Video Diffusion Models}
\label{subsec:vid}

As shown in Fig.~\ref{fig:pipeline1}, taking a single reference image $\mathbf{I}^{\text{ref}}$ as an example, we first obtain its point cloud, camera intrinsics and camera pose $\mathbf{C}^{\text{ref}}$ through the dense stereo model\cite{wang2024dust3r}. Subsequently, we can navigate the camera along a camera pose sequence $\mathbf{C} = \{\mathbf{C}^{0},...,\mathbf{C}^{L-1} \}$ that contains $\mathbf{C}^{\text{ref}}$  to render the point cloud and obtain a sequence of render results, denote as $\mathbf{P} = \{\mathbf{P}^{0},...,\mathbf{P}^{L-1} \}$. While the point cloud renders accurately represent view relationships, they are plagued by substantial occlusions, missing areas, and reduced visual fidelity. To enhance the quality of novel view rendering, our objective is to learn a conditional distribution $\bm{x} \sim p(\bm{x}~|~\mathbf{I}^{\text{ref}},\mathbf{P})$ that can produce high-quality novel views $\bm{x} = \{\bm{x}^{0},...,\bm{x}^{L-1}\}$ based on the point cloud renders $\mathbf{P}$ and the reference image(s) $\mathbf{I}^{\text{ref}}$. 
Motivated by the efficacy of video diffusion models\cite{xing2023dynamicrafter,pku_yuan,blattmann2023svd} in synthesizing high-quality and consistent videos, we learn this conditional distribution by training a video diffusion model conditioned on the point cloud renders and the reference image(s). As a result, the novel view synthesis process can be naturally modeled as the reverse process of a point-conditioned video diffusion model, expressed as $\bm{x} \sim p_\theta(\bm{x}~|~\mathbf{I}^{\text{ref}},\mathbf{P})$, where $\theta$ denotes the model parameters. 

The architecture of the point-conditioned video diffusion model is illustrated in Fig.~\ref{fig:pipeline1}. 
It inherits the LDM\cite{metzer2022latent} architecture, which primarily comprises a pair of VAE encoder $\mathcal{E}$ and decoder $\mathcal{D}$ for image compression, a video denoising U-Net with spatial layers followed by temporal layers for temporal-aware noise estimation, as well as a CLIP\cite{radford2021clip} image encoder for reference image understanding. 
We incorporate point cloud renders as conditional signals in the video denoising U-Net by encoding them using $\mathcal{E}$ and concatenating the resulting latent images with noise across the channel dimension.

To train the model, we create paired training data that includes both point cloud renders $\mathbf{P} = \{\mathbf{P}^{0},...,\mathbf{P}^{L-1} \}$ and the corresponding ground-truth reference images $\mathbf{I} = \{\mathbf{I}^{0},...,\mathbf{I}^{L-1} \}$. The point cloud renders are forced to traverse at least one ground-truth view, \emph{i.e.}, to include at least one ground-truth reference image at a random location among the $L$ frames. It helps the model better learn to transfer fine details from the reference image(s) to the point cloud renders and enables our model to flexibly handle arbitrary number of reference image(s). Following the approach of LDMs\cite{metzer2022latent}, we freeze the parameters of the VAE encoder $\mathcal{E}$ and decoder $\mathcal{D}$, and conduct the training process in the latent space. Specifically, we encode the training data pair $\mathbf{I} = \{\mathbf{I}^{0},...,\mathbf{I}^{L-1} \}$ and $\mathbf{P} = \{\mathbf{P}^{0},...,\mathbf{P}^{L-1} \}$ into the latent space, yielding the ground-truth latents $\bm{z} = \{\bm{z}^{0},...,\bm{z}^{L-1} \}$ and the condition signals $\hat{\bm{z}} = \{\hat{\bm{z}}^{0},...,\hat{\bm{z}}^{L-1} \}$ that will be concatenated channel-wise with the sampled noise. Subsequently,
the video denoising U-Net is optimized by the diffusion loss:
\begin{equation}
\min _{\theta} = \mathbb{E}_{t\sim\gU(0,1),\epsilon\sim\mathcal{N}(\bm{0},\mI)}[\|\epsilon_\theta(\bm{z}_t,t,\hat{\bm{z}},\mathbf{I}^{\text{ref}}) - \epsilon\|_2^2],
\end{equation}
where $\bm{z}_t = \alpha_t\bm{z}_0+\sigma_t\epsilon$.

During the inference process, we render a sequence of point cloud renders $\mathbf{P} = \{\mathbf{P}^{0},...,\mathbf{P}^{L-1} \}$ and replace the reference view render results with the corresponding reference image(s). Subsequently, we encode them into the latent space to obtain the latent images  $\hat{\bm{z}} = \{\hat{\bm{z}}^{0},...,\hat{\bm{z}}^{L-1} \}$,  sample noise $\epsilon\sim\mathcal{N}(\bm{0},\mI)$, then concatenate them channel-wise to construct the noisy latent. In addition to the latent space condition, we also pass the reference image(s) into the CLIP image\cite{radford2021clip} encoder, which will modulate the U-Net features through cross-attention for better 3D understanding. With the trained U-Net, the noisy latents are iteratively denoised into clean latents and then decoded into high-quality novel views $\bm{x} = \{\bm{x}^{0},...,\bm{x}^{L-1}\}$ using the VAE decoder $\mathcal{D}$.

\subsection{Iterative View Synthesis and Camera Trajectory Planning}
\label{subsec:NBV}
Existing video diffusion models encounter challenges in generating long videos with numerous frames. As the video length increases during inference, it results in decreased video stability and increased computational costs. Therefore, the trained ViewCrafter model may face challenges in generating longer videos to produce a larger view range.
To address this challenge, we adopt an iterative view synthesis approach. Specifically, given an initial point cloud established from the reference image(s), we navigate the camera from one of the reference views to a target camera pose to reveal occlusion and missing regions of the current point cloud. Subsequently, we can generate high-fidelity novel views using ViewCrafter and back-project the generated views to complete the point cloud. 
Through iteratively moving the camera, generating novel views, and updating the point cloud, we can ultimately obtain novel views with an extended view range and a more complete point cloud representation of the scene. 

\begin{algorithm}[tb]
\caption{Camera trajectory planning}
\label{algo}
\textbf{Input: }{Reference image(s) $\mathcal{I}_{\text{ref}}$, dense stereo model $\mathcal{D}(\cdot)$, point-conditioned video diffusion model $\mathcal{V}(\cdot)$, initial point cloud $\mathcal{P}_{\text{ref}}$, searching space $\mathcal{S}$, initial pose $\mathcal{C}_{\text{ref}}$, maximum predicted poses $N$, number of candidate poses $K$, utility function $\mathcal{F}(\cdot)$}

\begin{algorithmic}[1]
\STATE {\bfseries Initialize} Current point cloud $\mathcal{P}_{\text{curr}}\leftarrow\mathcal{P}_{\text{ref}}$, current camera pose $\mathcal{C}_{\text{curr}}\leftarrow\mathcal{C}_{\text{ref}}$, $step\leftarrow0$
\WHILE{$step \leq N$}
\STATE Uniformly sample $K$ candidate poses $\mathcal{C}_{\text{can}}={\{\mathcal{C}^{1}_{\text{can}}},...,{\mathcal{C}^{K}_{\text{can}}}\}$ from the searching space $S$ around the current pose $\mathcal{C}_{\text{curr}}$, initialize candidate mask set $\mathcal{M}_{\text{can}}=\{\}$  
\FOR{$\mathcal{C}$ in ${\{\mathcal{C}^{1}_{\text{can}}},...,{\mathcal{C}^{K}_{\text{can}}}\}$} 
\STATE $\mathcal{M}_{\mathcal{C}} = Render(\mathcal{P}_{\text{curr}},\mathcal{C})$
\STATE $\mathcal{M}_{\text{can}}.append(\mathcal{M}_{\mathcal{C}})$ 
\ENDFOR 
\STATE {
$\mathcal{C}_{\text{nbv}} = \mathop{\arg\max}\limits_{\mathcal{C} \in \mathcal{C}_{\text{can}}}\mathcal{F}(\mathcal{C})$
}
\vspace{.1cm}
\STATE $\mathcal{I}_{\text{nbv}} =  \mathcal{V}(interpolate(\mathcal{C}_{\text{curr}},\mathcal{C}_{\text{nbv}}),\mathcal{P}_{\text{curr}})$
\STATE $\mathcal{P}_{\text{curr}} \leftarrow  \mathcal{D}(\mathcal{I}_{\text{nbv}},\mathcal{P}_{\text{curr}})$
\STATE $\mathcal{C}_{\text{curr}} \leftarrow \mathcal{C}_{\text{nbv}}$
\STATE $step\leftarrow step+1$
\ENDWHILE
\STATE {\bfseries return} 
\end{algorithmic}
\end{algorithm}

In the iterative view synthesis process, the design of the camera trajectory significantly impacts the synthesis results. Methods like\cite{chung2023luciddreamer,gao2024cat3d} use predefined camera trajectories for scene generation, which overlooks the diverse geometry relationships presented in different scenes, resulting in significant occlusions.
To effectively reveal occlusions in the iterative view synthesis process and facilitate more complete scene generation, we designed a Next-Best-View (NBV)\cite{zeng2020viewsurvey,dhami2023pred,jin2023neu} -based camera trajectory planning algorithm, which enables the adaptive generation of camera trajectories tailored to handle various scene types.
The camera trajectory planning algorithm is illustrated in Algorithm.~\ref{algo}. 
Starting with the input reference image(s) $\mathcal{I}_{\text{ref}}$, we construct an initial point cloud $\mathcal{P}_{\text{ref}}$ using the dense stereo model\cite{wang2024dust3r}. Referring \cite{zeng2020viewsurvey, jin2023neu, isler2016informationgain, peralta2020next}, we opt for a forward-facing quarter-sphere with evenly distributed camera poses as the searching space, denoted as $\mathcal{S}$, and position it centrally at the origin of the point cloud's world coordinate system, setting the radius to the depth of the center pixel in the reference image.
The camera trajectory is initialized from one of the reference camera poses $\mathcal{C}_{\text{ref}}$. To predict the subsequent pose, we uniformly sample $K$ candidate camera poses $\mathcal{C}_{\text{can}}={\{\mathcal{C}^{1}_{\text{can}}},...,{\mathcal{C}^{K}_{\text{can}}}\}$ from the searching space surrounding the current camera pose $\mathcal{C}_{\text{curr}} = \mathcal{C}_{\text{ref}}$, then render a set of candidate masks $\mathcal{M}_{\text{can}}$ (where 1 signifies occlusion and missing regions, while 0 represents filled regions) from the current point cloud $\mathcal{P}_{\text{curr}}$. We then establish a utility function\cite{zeng2020viewsurvey} $\mathcal{F}(\cdot)$ to determine the optimal camera pose for the subsequent step, defined as: 
\begin{equation}
\mathcal{F}(\mathcal{C})=\left\{
\begin{aligned}
&\frac{\rm{sum}(\mathcal{M_{C}})}{W\times H}, \frac{\rm{sum}(\mathcal{M_{C}})}{W\times H} < \Theta \\
&1-\frac{\rm{sum}(\mathcal{M_{C}})}{W\times H}, \frac{\rm{sum}(\mathcal{M_{C}})}{W\times H} > \Theta, \\
\end{aligned}
\right.
\end{equation}
where $\mathcal{C}\in \mathcal{C}_{\text{can}}$, $\mathcal{M_{C}} \in \mathcal{M}_{\text{can}}$, and $\text{sum}(\mathcal{M_{C}}) = \sum_{u=0}^{W} \sum_{v=0}^{H}\mathcal{M_{C}}(u,v)$. The utility function helps identify a suitable camera pose that reveals an adequate area of occlusion and missing regions while avoiding poses that reveal excessively large holes deviating significantly from a threshold $\Theta$, which may affect ViewCrafter's generation capability.
Once the next best camera pose $\mathcal{C}_{\text{nbv}}$ is predicted, we interpolate a camera path between $\mathcal{C}_{\text{curr}}$ and $\mathcal{C}_{\text{nbv}}$, and then apply ViewCrafter along the path to generate a sequence of high-fidelity novel views. Subsequently, we back-project and align the generated novel view $\mathcal{I}_{\text{nbv}}$ onto the current point cloud $\mathcal{P}_{\text{curr}}$, and designate $\mathcal{C}_{\text{nbv}}$ as the new $\mathcal{C}_{\text{curr}}$, then repeat the aforementioned process until the predicted poses reach the predefined limitation $N$.
Through iteratively predicting camera poses, synthesizing novel views, and updating the point cloud, we can ultimately obtain a more complete point cloud representation of the scene.

\begin{figure*}[t]
  \centering
  \includegraphics[width=1.\textwidth]{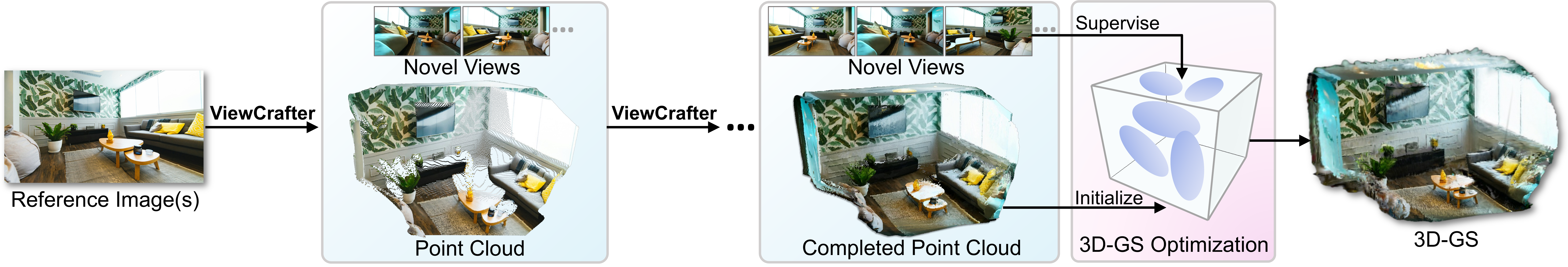}
  \caption{\textbf{Application of 3D-GS optimization.} To facilitate more consistent 3D-GS optimization, we leverage the iterative view synthesis strategy to progressively complete the initial point cloud and synthesize novel views using ViewCrafter. We then use the completed dense point cloud to initialize 3D-GS and employ the synthesized novel views to supervise 3D-GS training.}
\label{fig:pipeline2}
\end{figure*}
\subsection{Applications}
\label{subsec:3D-GS}
ViewCrafter can effectively produce accurate, consistent, and high-fidelity novel views from single or sparse inputs. 
Nevertheless, it faces challenges in providing immersive experiences due to the slow multi-step denoising process.
To achieve real-time rendering, we further delve into optimizing a 3D-GS\cite{kerbl20233dgs} representation from the results of our ViewCrafter.
To that aim, a direct approach involves concurrently running ViewCrafter multiple times on the initially built point cloud to generate multiple novel views and optimizing a 3D-GS from them.
However, this will lead to suboptimal optimization results, since the initial point cloud is incomplete and will introduce inconsistencies in occlusion regions among the generated view sequences at different times.

As illustrated in Fig.~\ref{fig:pipeline2}, to facilitate more consistent 3D-GS optimization, we leverage the aforementioned iterative view synthesis strategy to iteratively complete the initial point cloud and synthesize novel views using ViewCrafter, which not only provides consistent novel views as training data but also offers a strong geometry initialization for the 3D-GS\cite{kerbl20233dgs}.
During training, the center of each 3D Gaussian is initialized from the completed dense point cloud, and the attributes of each 3D Gaussian are optimized under the supervision of the synthesized novel views.
We simplify the 3D-GS optimization process by deprecating the densification, splitting, and opacity reset tricks\cite{shih20203d}, and reduce the overall optimization time into 2,000 iterations, which is considerably faster than the original 3D-GS training.

In addition to synthesizing novel views of real-world images, we also explore the application of combining ViewCrafter with creative text-to-image diffusion models for text-to-3D generation. This involves using a text-to-image diffusion model to generate a reference image from the provided text prompt, followed by employing ViewCrafter for novel view synthesis and 3D reconstruction.
\begin{figure*}[t]
  \centering
  \includegraphics[width=1.\textwidth]{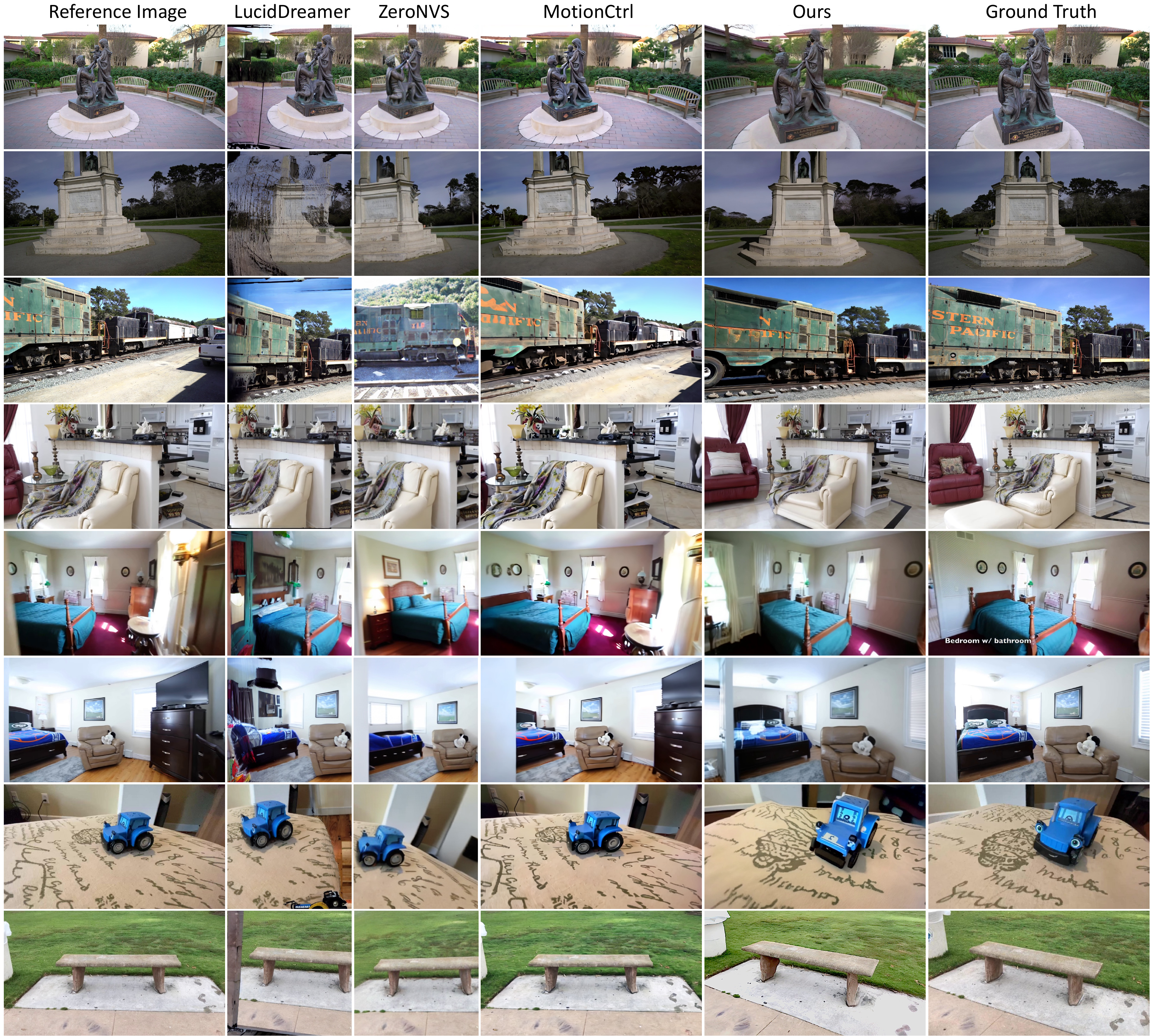}
  \caption{\textbf{Qualitative comparison of zero-shot novel view synthesis on Tanks-and-Temples\cite{Knapitsch2017tnt}, RealEstate10K\cite{zhou2018real10k} and CO3D\cite{reizenstein21co3d} dataset.}
  The reference images are displayed in the left-most column, and the ground truth novel views are located in the right-most column.
  }
\label{fig:compare_zero}
\end{figure*}
\begin{table*}[t]
\caption{\textbf{Quantitative comparison of zero-shot novel view synthesis on Tanks-and-Temples\cite{Knapitsch2017tnt}, RealEstate10K\cite{zhou2018real10k} and CO3D\cite{reizenstein21co3d} dataset.} Since ZeroNVS\cite{zeronvs} and LucidDreamer\cite{chung2023luciddreamer} can only handle square images, we crop the generated novel views from our method and MotionCtrl\cite{wang2023motionctrl} to align with them when computing the quantitative metrics. 
}
\label{tab:comp_zero}
\centering
\resizebox{0.9\textwidth}{!}{%
\huge
\begin{tabular}{lcccccccccccc}
\cmidrule[\heavyrulewidth]{1-13}
 \textbf{Dataset}
 & \multicolumn{6}{c}{\textbf{Easy set}} & \multicolumn{6}{c}{\textbf{Hard set}} \\
 \cmidrule(lr){2-7} \cmidrule(lr){8-13}
 Method & LPIPS $\downarrow$ & PSNR $\uparrow$ & SSIM $\uparrow$ & FID $\downarrow$ & $R_{\text{dist}}$ $\downarrow$ & $T_{\text{dist}}$ $\downarrow$ & LPIPS $\downarrow$ & PSNR $\uparrow$ & SSIM $\uparrow$ & FID $\downarrow$ & $R_{\text{dist}}$ $\downarrow$ & $T_{\text{dist}}$ $\downarrow$\\ \cmidrule{1-13}

\textbf{Tanks-and-Temples}\\
 LucidDreamer~\cite{chung2023luciddreamer}   & \cellcolor{tabthird}0.413 & 14.53 & 0.362 & \cellcolor{tabthird}42.32 & \cellcolor{tabsecond}6.137 & \cellcolor{tabsecond}5.695 & \cellcolor{tabthird}0.558 & 11.69 & 0.267 & 200.8 & 8.998 & 9.305 \\
  ZeroNVS~\cite{zeronvs}  & 0.482 & \cellcolor{tabthird}14.71 & \cellcolor{tabthird}0.380 & 74.60 & 8.810 & \cellcolor{tabthird}6.348 & 0.569 & \cellcolor{tabthird}12.05 & \cellcolor{tabthird}0.309 & \cellcolor{tabsecond}131.0 & \cellcolor{tabsecond}8.860 & \cellcolor{tabsecond}8.557\\
 MotionCtrl~\cite{wang2023motionctrl}  &  \cellcolor{tabsecond}0.400 &  \cellcolor{tabsecond}15.34 &  \cellcolor{tabsecond}0.427 &  \cellcolor{tabsecond}70.3 &  \cellcolor{tabthird}7.299 &  8.039 &  \cellcolor{tabsecond}0.473 &  \cellcolor{tabsecond}13.29 &  \cellcolor{tabsecond}0.384 &  \cellcolor{tabthird}196.8 &  \cellcolor{tabthird}9.801 &  \cellcolor{tabthird}9.112  \\
  ViewCrafter (ours) & \cellcolor{tabfirst}0.194 & \cellcolor{tabfirst}21.26 & \cellcolor{tabfirst}0.655 & \cellcolor{tabfirst}27.18 & \cellcolor{tabfirst}0.471 & \cellcolor{tabfirst}1.009 & \cellcolor{tabfirst}0.283 & \cellcolor{tabfirst}18.07 & \cellcolor{tabfirst}0.563 & \cellcolor{tabfirst}38.92 & \cellcolor{tabfirst}1.109 & \cellcolor{tabfirst}0.910 
  \\ \cmidrule{1-13}         
\textbf{RealEstate10K} \\
 LucidDreamer~\cite{chung2023luciddreamer}     & \cellcolor{tabsecond}0.315 & 16.35 & \cellcolor{tabthird}0.579 & 56.77 & \cellcolor{tabthird}5.821 & 10.02 & \cellcolor{tabthird}0.400 & 14.13 & 0.511 & \cellcolor{tabthird}71.43 & \cellcolor{tabsecond}7.990 & 10.85  \\
 ZeroNVS~\cite{zeronvs}    & 0.364 & \cellcolor{tabsecond}16.50 & 0.577 & \cellcolor{tabthird}96.18 & 6.370 & \cellcolor{tabthird}9.817 & 0.431 & \cellcolor{tabthird}14.24 & \cellcolor{tabthird}0.535 & 105.8 & 8.562 & \cellcolor{tabthird}10.31 \\ 
 MotionCtrl~\cite{wang2023motionctrl}  & \cellcolor{tabthird}0.341 & \cellcolor{tabthird}16.31 & \cellcolor{tabsecond}0.604 & \cellcolor{tabsecond}89.90 & \cellcolor{tabsecond}4.236 & \cellcolor{tabsecond}9.091 & \cellcolor{tabsecond}0.386 & \cellcolor{tabsecond}16.29 & \cellcolor{tabsecond}0.587 & \cellcolor{tabsecond}70.02 & \cellcolor{tabthird}8.084 & \cellcolor{tabsecond}9.295 \\
  ViewCrafter (ours) & \cellcolor{tabfirst}0.145 & \cellcolor{tabfirst}21.81 & \cellcolor{tabfirst}0.796 & \cellcolor{tabfirst}33.09 & \cellcolor{tabfirst}0.380 & \cellcolor{tabfirst}2.888 & \cellcolor{tabfirst}0.178 & \cellcolor{tabfirst}22.04 & \cellcolor{tabfirst}0.798 & \cellcolor{tabfirst}24.89 & \cellcolor{tabfirst}1.098 & \cellcolor{tabfirst}2.867
  \\ \cmidrule{1-13}                      
\textbf{CO3D} \\
 LucidDreamer~\cite{chung2023luciddreamer}     & \cellcolor{tabthird}0.429 & 15.11 & 0.451 & \cellcolor{tabthird}78.87 & 12.90 & \cellcolor{tabthird}6.665 & \cellcolor{tabthird}0.517 & 12.69 & 0.374 & 157.8 & \cellcolor{tabthird}16.43 & \cellcolor{tabthird}8.301  \\ 
 ZeroNVS~\cite{zeronvs}    & 0.467 & \cellcolor{tabthird}15.15 & \cellcolor{tabthird}0.463 & 93.84 & \cellcolor{tabthird}15.44 & 8.872 & 0.524 &  \cellcolor{tabthird}13.31 & \cellcolor{tabthird} 0.426 & \cellcolor{tabthird}143.2 &  \cellcolor{tabsecond}15.02 &  10.22\\
 MotionCtrl~\cite{wang2023motionctrl}  & \cellcolor{tabsecond}0.393 & \cellcolor{tabsecond}16.87 & \cellcolor{tabsecond}0.529 & \cellcolor{tabsecond}69.18 & \cellcolor{tabsecond}16.87 & \cellcolor{tabsecond}5.131 & \cellcolor{tabsecond}0.443 & \cellcolor{tabsecond}15.46 & \cellcolor{tabsecond}0.502 & \cellcolor{tabsecond}112.7 & 18.81 & \cellcolor{tabsecond}5.575  \\
 ViewCrafter (ours) & \cellcolor{tabfirst}0.243 & \cellcolor{tabfirst}21.38 & \cellcolor{tabfirst}0.687 & \cellcolor{tabfirst}24.63 & \cellcolor{tabfirst}2.175 & \cellcolor{tabfirst}1.033 & \cellcolor{tabfirst}0.324 & \cellcolor{tabfirst}18.96 & \cellcolor{tabfirst}0.641 & \cellcolor{tabfirst}36.96 & \cellcolor{tabfirst}2.849 & \cellcolor{tabfirst}1.480\\
 \cmidrule[\heavyrulewidth]{1-13}
\end{tabular}
}
\end{table*}
\section{Experiments}
In this section, we begin with an illustration of the implementation details in Section.~\ref{subsec:implementation}, followed by a comparison of zero-shot novel view synthesis in Section.~\ref{subsec:comp_zero} and scene reconstruction in Section.~\ref{subsec:reconstruction}. Subsequently, we conduct ablation studies to evaluate design choices and training strategies in Section.~\ref{subsec:ablation}. Finally, we present the results of text-to-3D generation in Section.~\ref{subsec:application}.
\subsection{Implementation Details}\label{subsec:implementation}
We employ a progressive training strategy. In the first stage, we train the ViewCrafter model at a resolution of 320 $\times$ 512, with the frame length set to 25. The entire video denoising U-Net is trained for 50,000 iterations using a learning rate of 5 $\times$ 10$^{-5}$ and a mini-batch size of 16. 
In the second stage, we fine-tune the spatial layers (\emph{i.e.}, 2D Conv and spatial attention layers) of the video denoising U-Net at a resolution of 576$\times$1024 for high-resolution adaptation, with 5,000 iterations on a learning rate of 1$\times$10$^{-5}$ and a valid mini-bach size of 16. 
Our model was trained on a mixed dataset consisting of DL3DV\cite{ling2024dl3dv} and RealEstate10K\cite{zhou2018real10k}. We divide the video data into video clips, each containing 25 frames. To generate the conditional signals, specifically the point cloud renders, we process the video clips using DUSt3R\cite{wang2024dust3r} to obtain the camera trajectory of the video clips and the globally aligned point clouds of each video frame. Then, we randomly select the constructed point cloud of the video frames and render it along the estimated camera trajectory using Pytorch3D\cite{ravi2020pytorch3d}. In total, we generate 632,152 video pairs as training data.
During inference, we adopt DDIM sampler\cite{song2021denoising} with classifier-free guidance\cite{ho2022classifier}.

\subsection{Zero-shot Novel View Synthesis Comparison}\label{subsec:comp_zero}
\noindent\textbf{Datasets and evaluation metrics.}
In our study, we employ three real-world datasets of different scales as our zero-shot novel view synthesis evaluation benchmark, which includes the CO3D\cite{reizenstein21co3d} dataset, the RealEstate10K\cite{zhou2018real10k} dataset, and the Tanks-and-Temples\cite{Knapitsch2017tnt} dataset.
For CO3D\cite{reizenstein21co3d} consisting of object-centric scenes, we evaluate on 10 scenes. RealEstate10K\cite{zhou2018real10k} comprises video clips of indoor scenes, we adopt 10 scenes from its test set for evaluation. For Tanks-and-Temples\cite{Knapitsch2017tnt} containing large-scale outdoor and indoor scenes, we use all of the 9 scenes.
For all benchmarks, we extract frames from the original captured videos and create two types of test sets by applying different sampling rates to the original video. The easy test set is generated using a small frame sampling stride, characterized by slow camera motions and limited view ranges. In contrast, the hard test set is produced with a large sampling stride, featuring rapid camera motions and large view ranges.

We employ PSNR, SSIM\cite{wang2004image}, LPIPS\cite{zhang2018unreasonable}, and FID\cite{heusel2017gans} as the evaluation metrics for assessing image quality. Among these, PSNR is a traditional metric used to compare image similarity.  SSIM\cite{wang2004image} and LPIPS\cite{zhang2018unreasonable} are designed to evaluate the structural and perceptual similarity between the generated images and the ground truth images, as these metrics are specifically designed to align more closely with human perceptual judgment. Referring to\cite{wiles2020synsin,rombach2021geometryfree,rockwell2021pixelsynth}, we further integrate FID into our evaluation process for assessing the quality of synthesized views, which proves particularly efficacious when evaluating the hard test set that contains a significant number of missing and occlusion regions. 
Additionally, to evaluate the pose accuracy of the generated novel views, we estimate the camera poses of the generated novel views to compare with the ground truth camera poses.  Following\cite{he2024cameractrl}, we transform the camera coordinate of the estimated poses to be relative to the first frame, and normalize the translation scale using the furthest frame. We then calculate the rotation distance ($R_{\text{dist}}$) in comparison to the ground truth rotation matrices of each generated novel view sequence, expressed as:
\begin{equation}
    \label{eq:r_dist}
    R_{\text{dist}} = \sum_{i=1}^n \arccos({\frac{\text{tr}(\mathbf{R}_{\text{gen}}^{i} \mathbf{R}_{\text{gt}}^{i\mathrm{T}}) - 1}{2}}),
\end{equation}
where $\mathbf{R}_{\text{gt}}^{i}$ and $\mathbf{R}_{\text{gen}}^{i}$ denote the ground truth rotation matrix and generated rotation matrix, and we sum the distance of all frames as the final results.
We also compute the translation distance ($T_{\text{dist}}$), expressed as:
 \begin{equation}
     \label{eq:t_dist}
     T_{\text{dist}} = \sum_{i=1}^n \|\mathbf{T}_{\text{gt}}^{i} - \mathbf{T}_{\text{gen}}^{i} \|_2.
 \end{equation}
Notably, since COLMAP\cite{schoenberger2016sfm} is sensitive to inconsistent features and prone to fail to extract poses from the generated novel views, we instead use DUSt3R\cite{wang2024dust3r} for more robust pose estimation.

\noindent\textbf{Comparison baselines.}
As a diffusion-based generalizable novel view synthesis framework, we compare our method with three diffusion-based baselines: ZeroNVS\cite{zeronvs}, MotionCtrl\cite{wang2023motionctrl} and LucidDreamer\cite{chung2023luciddreamer}. Specifically, ZeroNVS\cite{zeronvs} is finetuned from Zero-1-to-3\cite{liu2023zero} and can generate novel views conditioned on a reference image and the relative camera pose. The camera pose is processed as CLIP\cite{radford2021clip} text embedding and injected into the diffusion U-Net via cross-attention. MotionCtrl\cite{wang2023motionctrl} is a camera-conditioned video diffusion model finetuned from SVD\cite{blattmann2023svd}. It can generate consistent novel views from the conditioned reference image and the relative camera pose sequences, which are also processed as high-level embedding and injected into the video diffusion U-Net through cross-attention. LucidDreamer\cite{chung2023luciddreamer} utilizes depth-based warping to synthesize novel views, and employs a pretrained diffusion-based inpainting model\cite{Rombach2022sd2} to inpaint missing regions in the novel views. 
For the zero-shot novel view synthesis comparison, we use a single reference image as input for all baselines and our method, since the baselines are only capable of performing single-view novel view synthesis. 

\begin{figure*}[t]
  \centering
  \includegraphics[width=1.\textwidth]{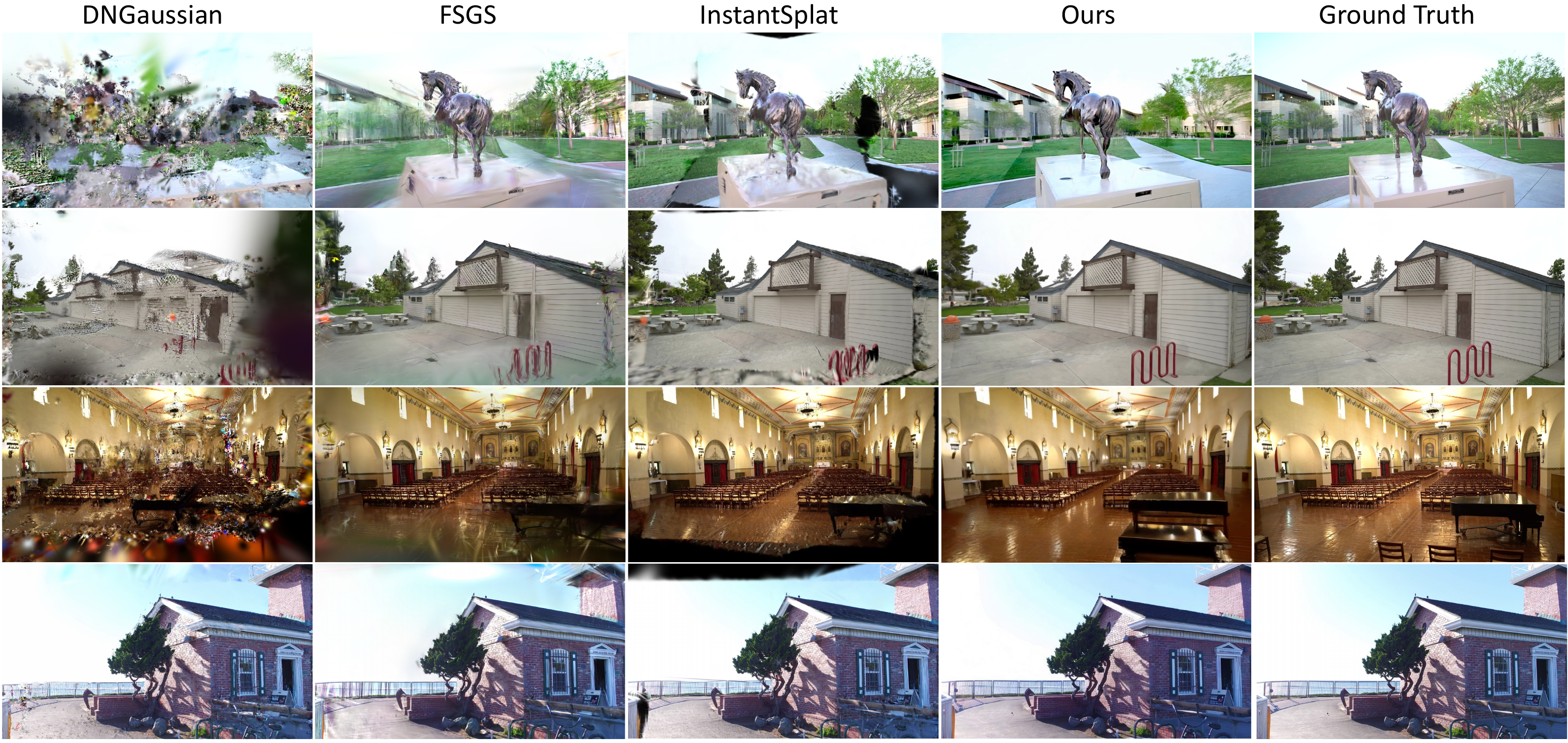}
  \caption{\textbf{Qualitative comparison 
  of scene reconstruction on Tanks-and-Temples\cite{Knapitsch2017tnt} dataset.} We train each scene using 2 ground truth training images, and render novel views to compare with the ground truth novel view (right-most column).
  }
\label{fig:compare_gs}
\end{figure*}
\noindent\textbf{Qualitative comparison.}
The qualitative results are presented in Fig.~\ref{fig:compare_zero}, where the reference images are displayed in the left-most column, and the ground truth novel views are located in the right-most column. The results of LucidDreamer\cite{chung2023luciddreamer} exhibit severe artifacts, since it uses depth-based warping for generating novel views,  which is particularly problematic when handling in-the-wild images with unknown camera intrinsic, leading to inaccurate novel views. Moreover, it employs an off-the-shelf inpainting model\cite{Rombach2022sd2} to refine the warped results, which tends to introduce inconsistencies between the original and inpainted content. Novel views generated by ZeroNVS\cite{zeronvs} also exhibit relatively low quality and poor accuracy; the primary reason is that ZeroNVS introduces the camera pose condition into diffusion models through text embedding, which fails to provide precise control over the generation of novel views, leading to sub-optimal results. Similarly, although MotionCtrl\cite{wang2023motionctrl} can produce novel views with better fidelity, it falls short in generating novel views that precisely align with the given camera conditions. This is because MotionCtrl also adopts a high-level camera embedding to control camera pose, leading to less accurate novel view synthesis. 
In comparison, our method incorporates explicit point cloud prior and video diffusion model, the results demonstrate the superiority of our method in terms of both pose control accuracy and the overall quality of the generated novel views. 

\noindent\textbf{Quantitative comparison.}
The quantitative comparison results are reported in Table.~\ref{tab:comp_zero}. Since ZeroNVS\cite{zeronvs} and LucidDreamer\cite{chung2023luciddreamer} can only handle squared images, we crop the generated novel views of our method and MotionCtrl\cite{wang2023motionctrl} to align with ZeroNVS and LucidDreamer when computing the quantitative metrics.
In terms of image quality, it can be observed that our approach consistently outperforms the baselines in all the metrics. Specifically, the higher PSNR and SSIM values indicate that our method maintains better image quality and similarity to the ground truth. The lower LPIPS score further demonstrates that our approach generates more perceptually accurate images, while the significantly improved FID score suggests that our method captures the underlying distribution of the data more effectively. In terms of pose accuracy, the reduced $R_{\text{dist}}$ and $T_{\text{dist}}$ demonstrate the effectiveness of our model design, which enables more accurate pose control in novel view synthesis.

\subsection{Scene Reconstruction Comparison}
\label{subsec:reconstruction}
\noindent\textbf{Datasets and evaluation metrics.} 
In the scene reconstruction comparison, we use 6 scenes from the Tanks-and-Temples dataset\cite{Knapitsch2017tnt} for evaluation. We create a challenging sparse-view benchmark that contains only 2 ground truth training images for each scene, and use 12 views for evaluation.
We employ PSNR, SSIM\cite{wang2004image}, and LPIPS\cite{zhang2018unreasonable} as the evaluation metrics for image quality assessment. 

\noindent\textbf{Comparison baselines.}
We compare our method with three 3D-GS representation-based sparse view reconstruction methods: DNGaussian\cite{li2024dngaussian}, FSGS\cite{zhu2023fsgs} and InstantSplat\cite{fan2024instantsplat}. 
Specifically, DNGaussian\cite{zeronvs} and FSGS\cite{zhu2023fsgs} utilize point cloud produced by COLAMP\cite{schoenberger2016sfm} as initialization, and leverage both image supervision and depth regularization for sparse view reconstruction. InstantSplat\cite{fan2024instantsplat} explores utilizing point cloud produced by DUSt3R\cite{wang2024dust3r} as initialization, which enables efficient 3D-GS training from sparse images.

\begin{table}[t]
\caption{\textbf{Quantitative comparison of scene reconstruction on Tanks-and-Temples\cite{Knapitsch2017tnt}.} We use 2 ground truth training images for each scene, and adopt 12 views for evaluation.
}
\label{tab:comp_gs}
\centering
\resizebox{0.45\textwidth}{!}{%
\setlength{\tabcolsep}{8 mm}
\huge
\begin{tabular}{lccccc}
\cmidrule[\heavyrulewidth]{1-4}
 \textbf{Method} & LPIPS $\downarrow$ & PSNR $\uparrow$ & SSIM $\uparrow$   \\ \cmidrule{1-4}
 DNGausian~\cite{li2024dngaussian}  &  \cellcolor{tabthird}0.331 &  15.47 &  0.541    \\
 FSGS~\cite{zhu2023fsgs}  &  0.364 &  \cellcolor{tabthird}17.53 &  \cellcolor{tabthird}0.558   \\
 InstantSplat~\cite{fan2024instantsplat}  &  \cellcolor{tabsecond}0.275 &  \cellcolor{tabsecond}18.61 &  \cellcolor{tabsecond}0.614     \\
  ViewCrafter (ours) & \cellcolor{tabfirst}0.245 & \cellcolor{tabfirst}21.50 & \cellcolor{tabfirst}0.692 
  \\ \cmidrule{1-4}         
 \cmidrule[\heavyrulewidth]{1-4}
\end{tabular}
}
\end{table}
\begin{figure*}[t]
  \centering
  \includegraphics[width=1.\textwidth]{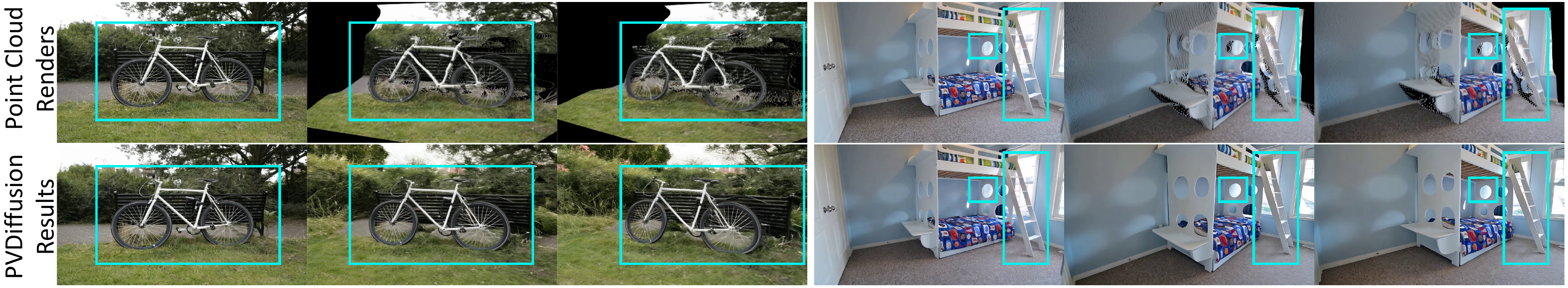}
  \caption{\textbf{Robustness for point cloud condition.} We show the point cloud render results and the corresponding novel views generated by ViewCrafter in the top and bottom rows, respectively (Best viewed with zoom-in). 
  }
\label{fig:ablate_robust}
\end{figure*}
\begin{figure*}[t]
  \centering
  \includegraphics[width=.95\textwidth]{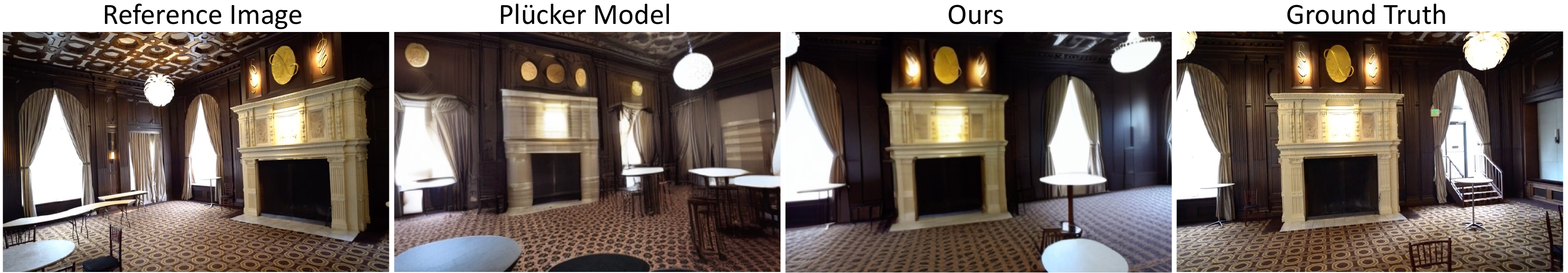}
  \caption{\textbf{Qualitative ablation of pose condition strategy.} 
We make the architecture and training strategy of the Plücker model to be identical to those of ViewCrafter, with the exception of the condition signal.
  }
\label{fig:ablate_plk}
\end{figure*}
\begin{figure}[t]
  \centering
  \includegraphics[width=.49\textwidth]{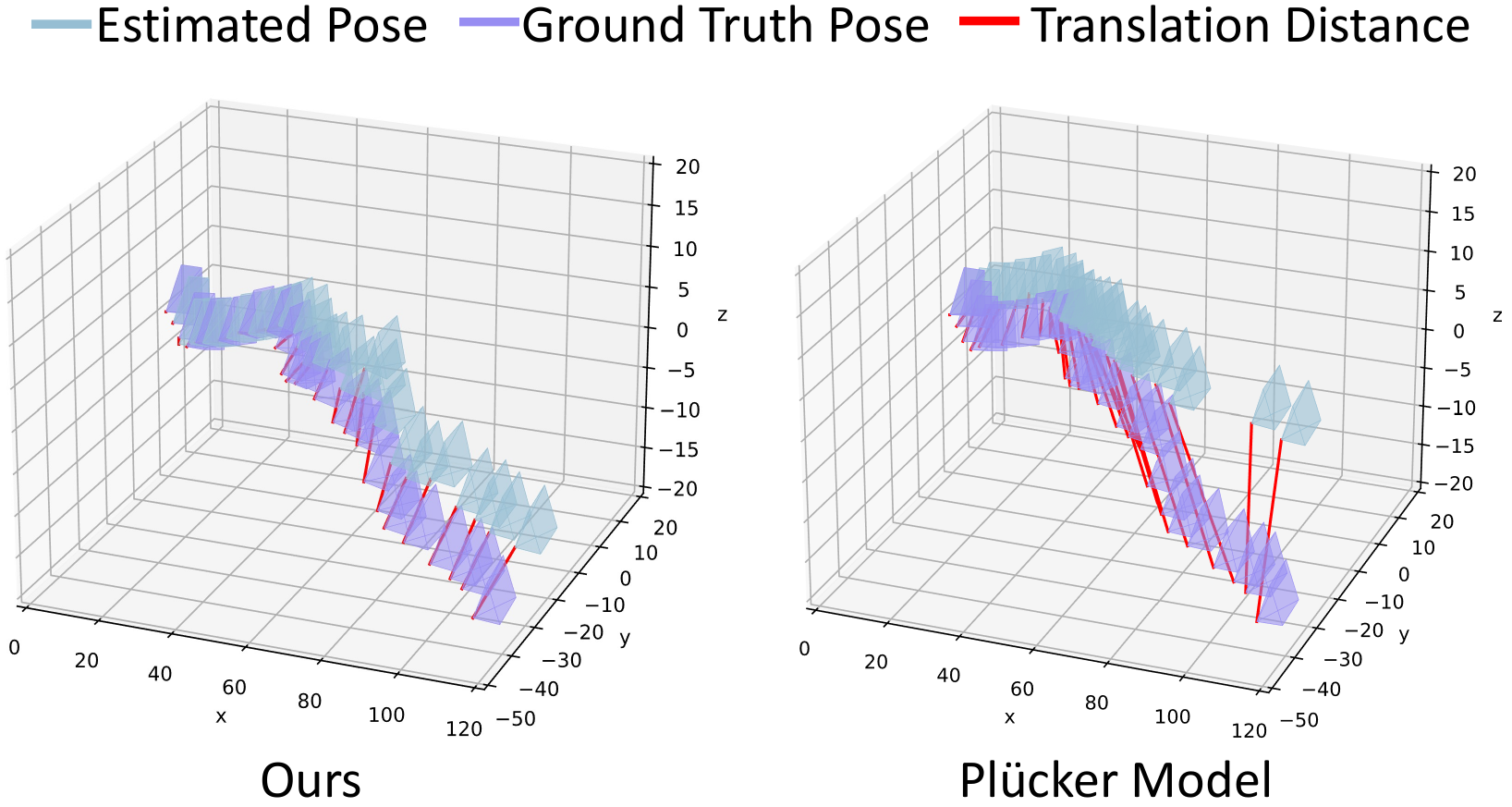}
  \caption{\textbf{Visualization of pose accuracy.} We compare the alignment level between the ground truth camera poses and the poses estimated from the generated novel views of ViewCrafter and the Plücker model. 
  }
\label{fig:ablate_pose}
\end{figure}
\begin{figure}[t]
  \centering
  \includegraphics[width=.49\textwidth]{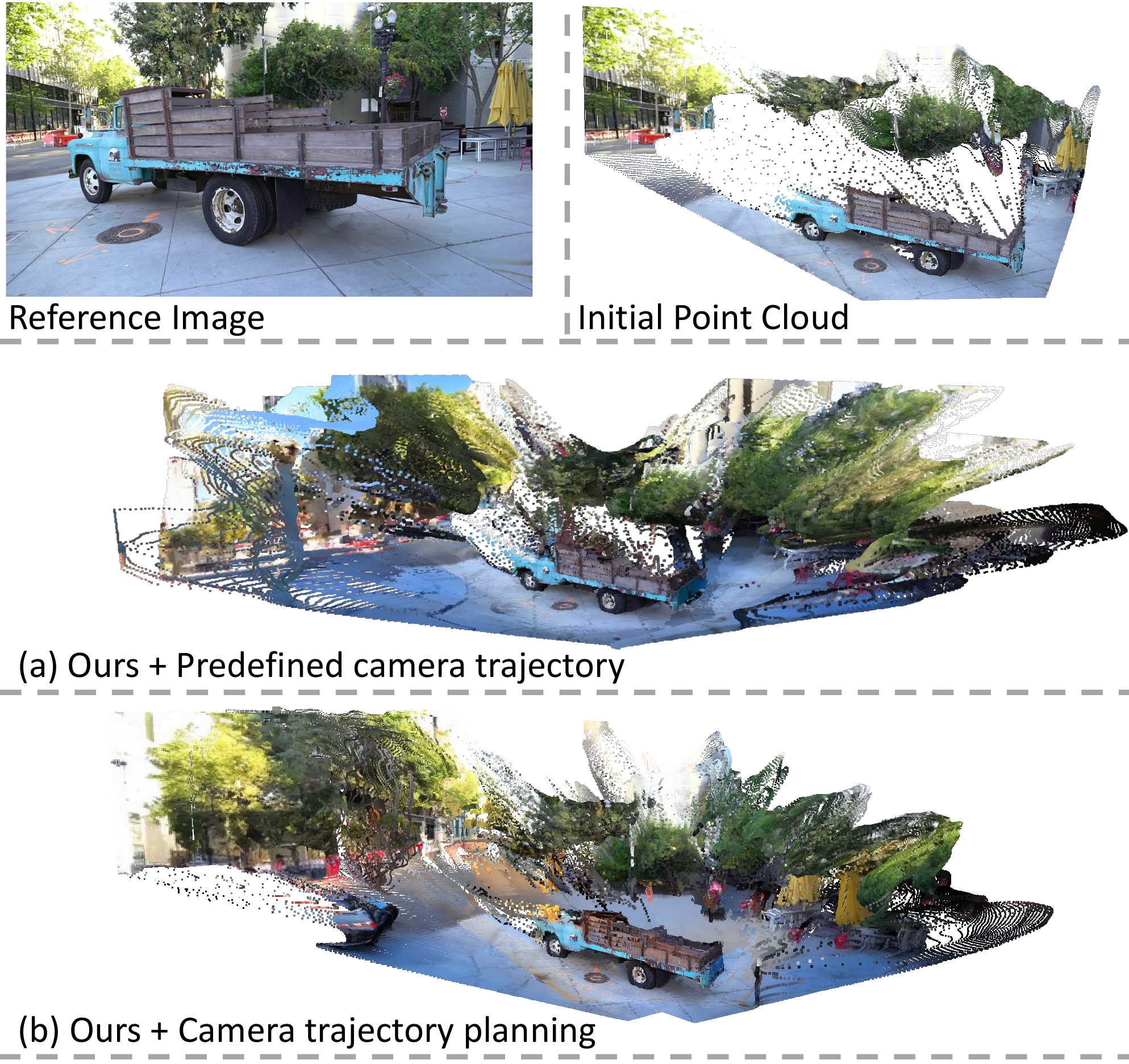}
  \caption{\textbf{Ablation study on the camera trajectory planning.} The point cloud reconstructed using the predefined camera trajectory fails to effectively complete the occlusion region. In contrast, the point cloud generated through our camera trajectory planning algorithm reveals the occlusion region of the scene more effectively, enhancing the overall reconstruction quality of the point cloud.
  }
\label{fig:ablate_traj}
\end{figure}
\noindent\textbf{Qualitative comparison.}
The qualitative comparison results are presented in Fig.~\ref{fig:compare_gs}. It can be observed the results from DNGaussian\cite{li2024dngaussian} exhibit significant artifacts. Similarly, results from FSGS\cite{zhu2023fsgs} show artifacts when viewed from novel views that deviate from the ground truth training images. Although InstantSplat\cite{fan2024instantsplat} utilizes DUSt3R\cite{wang2024dust3r} for point cloud initialization, which better preserves details from the ground truth training images, it fails to recover occlusion regions due to its omission of the densification process\cite{kerbl20233dgs}, resulting in severe holes under novel views. In comparison, our method leverages the priors from video diffusion models, enabling the generation of high-fidelity novel views given only 2 ground truth training images.

\noindent\textbf{Quantitative comparison.}
The quantitative comparison results are presented in Table.~\ref{tab:comp_gs}. It can be observed that our approach consistently outperforms the comparison baselines in all the metrics, further validating the effectiveness of our method in scene reconstruction from sparse views.

\begin{table}[t]
\caption{\textbf{Ablation study of the pose condition strategy.} 
Except for the condition signal, the architecture and training strategy of the Plücker model are identical to those of ViewCrafter.
}
\label{tab:ablate_plk}
\centering
\resizebox{0.47\textwidth}{!}{%
\huge
\begin{tabular}{lcccccc}
\cmidrule[\heavyrulewidth]{1-7}
 \textbf{Method} & LPIPS $\downarrow$ & PSNR $\uparrow$ & SSIM $\uparrow$ & FID $\downarrow$ & $R_{\text{dist}}$ $\downarrow$ & $T_{\text{dist}}$ $\downarrow$ \\ \cmidrule{1-7}
 Plücker model  &  0.370 &  17.51 &  0.546 &  49.33 &  2.688 &  2.570   \\
  Ours & \cellcolor{tabfirst}0.270 & \cellcolor{tabfirst}20.25 & \cellcolor{tabfirst}0.649 & \cellcolor{tabfirst}38.17 & \cellcolor{tabfirst}0.552 & \cellcolor{tabfirst}0.983
  \\ \cmidrule{1-7}         
 \cmidrule[\heavyrulewidth]{1-7}
\end{tabular}
}
\end{table}
\begin{table}[t]
\caption{\textbf{Quantitative comparison of different training paradigms.} 
We analyze the effectiveness of training both the spatial and temporal layers of the video denoising U-Net, as well as the benefits of the progressive training strategy and inference with more frames.
)
}
\label{tab:ablate_train}
\centering
\resizebox{0.47\textwidth}{!}{%
\huge
\begin{tabular}{lcccc}
\cmidrule[\heavyrulewidth]{1-5}
 \textbf{Traing paradigm} & LPIPS $\downarrow$ & PSNR $\uparrow$ & SSIM $\uparrow$ & FID $\downarrow$  \\ \cmidrule{1-5}
 Only train spatial layers  &  0.301 &  18.82 &  0.595 &  42.30    \\
 Directly train on 576$\times$1024  &  0.314 &  18.55 &  0.582 &  41.01    \\
 16 frames model  &  0.289 &  19.07 &  0.610 &  38.43    \\
  Ours & \cellcolor{tabfirst}0.280 & \cellcolor{tabfirst}19.52 & \cellcolor{tabfirst}0.615 & \cellcolor{tabfirst}37.77 
  \\ \cmidrule{1-5}         
 \cmidrule[\heavyrulewidth]{1-5}
\end{tabular}
}
\end{table}
\begin{figure*}[t]
  \centering
  \includegraphics[width=1.\textwidth]{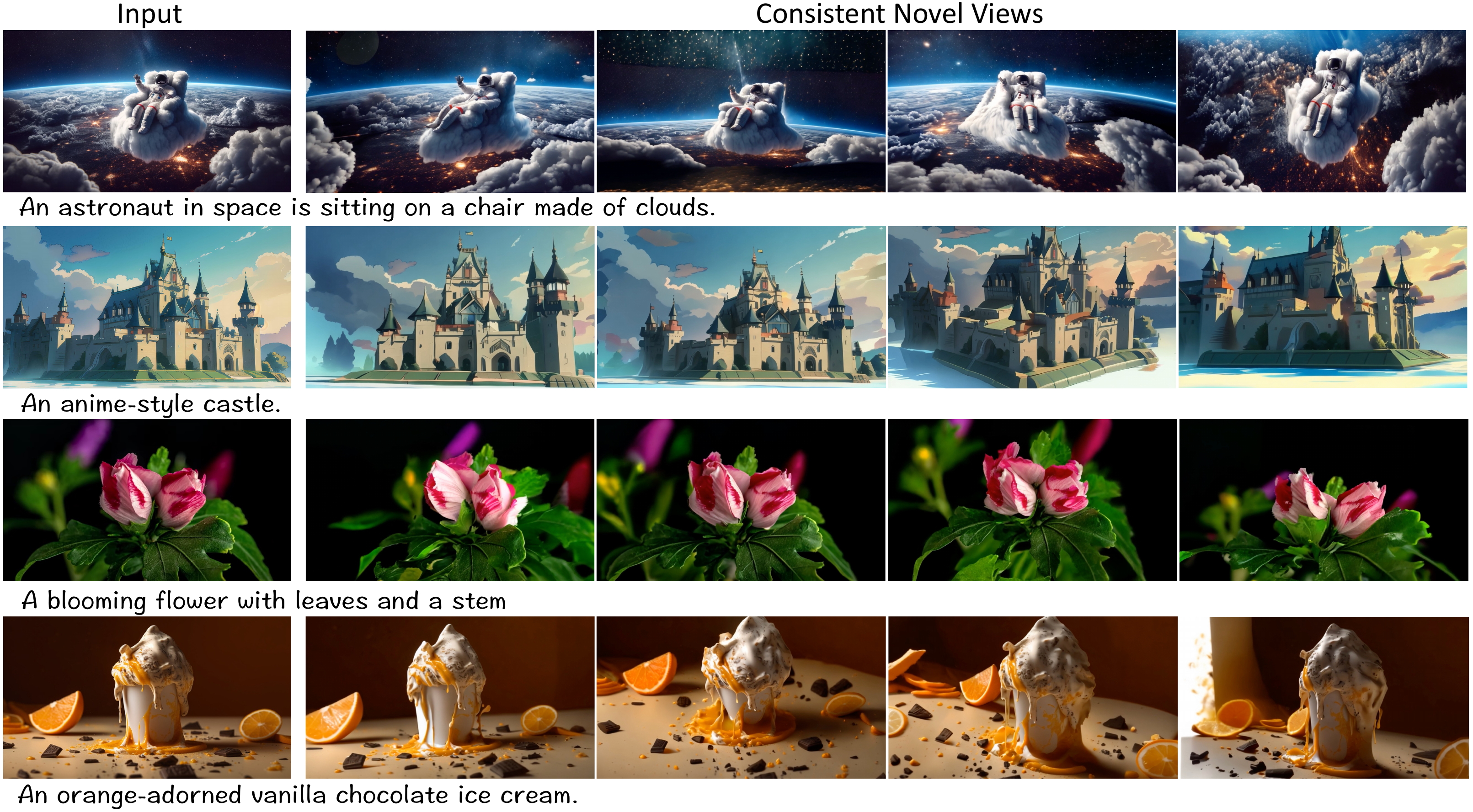}
  \caption{\textbf{Text-to-3D generation results.} The leftmost column displays the input text prompt and the corresponding image generated by T2I model, while the subsequent columns show the consistent novel views produced by our ViewCrafter based on the generated image.}
\label{fig:t23}
\end{figure*}
\subsection{Ablation Study}
\label{subsec:ablation}

\noindent\textbf{Discussion on pose condition strategy.}
In our method, we utilize point cloud renders as an explicit condition for the video diffusion model, enabling highly accurate pose control for novel view synthesis. Some concurrent works\cite{xu2024camco,bahmani2024vd3d} adopt Plücker coordinates\cite{jia2020plucker} as pose condition for pose-controllable video generation. To compare the pose accuracy between our point cloud-based pose condition strategy and the Plücker coordinate-based pose condition strategy, we train a Plücker coordinate-conditioned video diffusion model (Plücker model for short) that accepts Plücker coordinates as conditions for synthesizing novel views.
The Plücker coordinate describes per-pixel motion; For a given RGB frame and its camera pose, its Plücker coordinate shares the same spatial size with the RGB frame and comprises 6 channels for each pixel location. Given a pose sequence, we resize the corresponding Plücker coordinates to align with the size of the latent space, and concatenate them with noise along the channel dimension. Except for the pose condition strategy, we maintain the rest of the architecture of the Plücker model to be identical to ViewCrafter, and train the model at $320 \times 512$ resolution, following the training details reported in Section.~\ref{subsec:implementation}. 
We conduct a zero-shot novel view synthesis comparison between ViewCrafter ($320 \times 512$ resolution) and the Plücker model. The qualitative and quantitative results are shown in Fig.~\ref{fig:ablate_plk} and Table.~\ref{tab:ablate_plk}, which demonstrates that the point cloud-based pose condition strategy employed in ViewCrafter achieves more accurate pose control in novel view synthesis. 
We also observed that the Plücker model prone to ignore the high-frequency movements of the camera. Fig.~\ref{fig:ablate_pose} presents an example, where we compare the alignment level between the ground truth camera poses and the poses estimated from the generated novel views. The results show that the poses estimated from the novel views generated by ViewCrafter align more closely with the ground truth poses, further demonstrating the effectiveness of our point cloud-based pose condition strategy.

\noindent\textbf{Robustness for point cloud condition.}
ViewCrafter utilizes point cloud render results as conditions, enabling highly accurate pose control. However, these renders may contain artifacts and geometric distortions. Fig.~\ref{fig:ablate_robust} presents an example: the first row shows that the conditioned point cloud renders exhibit occlusions and missing regions, as well as geometric distortions along the boundary of the foreground. The second row displays the corresponding novel views produced by ViewCrafter, demonstrating its ability to fill in the holes and correct the inaccurate geometry. This demonstrate that ViewCrafter has developed a comprehensive understanding of the 3D world, allowing it to generate high-quality novel views from imperfect conditional information and exhibit robustness to the point cloud conditions.

\noindent\textbf{Ablation on training paradigm.}
We examine the effectiveness of the adopted training paradigm in this ablation study.
To evaluate the choice of training which module of the video denoising U-Net, we compare the novel view synthesis quality of training only the spatial layers and training both the spatial and temporal layers (Ours), the results are shown in the first row of Table.~\ref{tab:ablate_train}.
To assess the importance of progressive training, we provide a comparison between directly training the model at $576 \times 1024$ resolution and training the model using the progressive training strategy (Ours), the results are reported in the second row of Table.~\ref{tab:ablate_train}.
Additionally, we have observed that the inference length of ViewCrafter influences the quality of novel view synthesis. Specifically, for the same range of view change, inference with more frames improves the temporal consistency of the generated frames. To balance the computational cost and synthesis quality, we train two models: a base model that infers 16 frames and a stronger model (Ours) that infers 25 frames. We present a comparison between the two models in the third row of Table.~\ref{tab:ablate_train}. The results above showcase the effectiveness of the implemented training paradigm.

\noindent\textbf{Ablation on camera trajectory planning.}
In this ablation study, we assess the effectiveness of our proposed camera trajectory planning algorithm for revealing occlusions and completing point clouds. An example is presented in Fig.~\ref{fig:ablate_traj}, where we compare the point cloud generated by iterative view synthesis using a predefined circular camera trajectory (similar as\cite{chung2023luciddreamer}) with that produced using our camera trajectory planning algorithm. Given a reference image and an initial point cloud, we adopt a quarter-sphere searching space centered at the origin of the initial point cloud's world coordinate system, with the radius set to the depth of the center pixel of the reference image. We begin with exploring the left half area of the searching space, with the parameters for camera trajectory planning set to $N=3$, $K=5$, and $\Theta=0.6$. Accordingly, the circular camera trajectory is set as uniformly moving the camera three times from the reference pose to the left direction of the searching space, with each movement measuring 20 degrees. Subsequently, we apply the same parameters to explore the right half of the searching space. The final generated point cloud are shown in Fig.~\ref{fig:ablate_traj}(a) and Fig.~\ref{fig:ablate_traj}(b).
In Fig.~\ref{fig:ablate_traj}(a), It can be observed that the point cloud reconstructed using the predefined circular camera trajectory results in ineffective completion of the occlusion region. 
In comparison, Fig.~\ref{fig:ablate_traj}(b) presents the reconstruction using our camera trajectory planning algorithm. The more complete reconstruction results demonstrate that it can effectively reveal occlusion regions of the scene, improving the overall scene reconstruction quality.

\subsection{Text-to-3D Generation}
\label{subsec:application}
In addition to synthesizing novel views of real-world images, we also explore the application of combining our framework with creative text-to-image (T2I) diffusion models for text-to-3D generation. To accomplish this, given a text prompt, we first adopt T2I models to generate a corresponding image, then apply ViewCrafter to synthesize consistent novel views. The results are shown in Fig.~\ref{fig:t23}.

\section{Conclusion and limitation}

This work presents ViewCrafter, a novel view synthesis framework that combines video diffusion models and point cloud priors for high-fidelity and accurate novel view synthesis. Our method overcomes the limitations of existing approaches by providing generalization ability for various scene types and adaptability for both single and sparse image inputs, while maintaining consistency and accuracy in the quality of novel views. 
Additionally, we introduce an iterative view synthesis method and an adaptive camera trajectory planning procedure that facilitate long-range novel view synthesis and automatic camera trajectory generation for diverse scenes.
Beyond novel view synthesis, we explore the efficient optimization of a 3D-GS representation for real-time, high frame-rate novel view rendering, and adapting our framework for text-to-3D generation.

\noindent\textbf{Limitations.} Despite its advantages, our method still has several limitations. Firstly, it may encounter challenges in synthesizing novel views with a very large view range given limited 3D clues, such as generating a front-view image from only a back-view image. Additionally, we leverage point clouds as an explicit prior and have validated the robustness of our method for low-quality point clouds. However, challenges may still persist in scenes where the conditioned point clouds are significantly inaccurate. Furthermore, as a video diffusion model, our method necessitates multi-step denoising during the inference process, which requires a relatively higher computing cost.

\end{document}